%% file: emnlp2022.tex
\pdfoutput=1

\documentclass[11pt]{article}

\usepackage[]{EMNLP2022}

\usepackage{times}
\usepackage{latexsym}

\usepackage[T1]{fontenc}

\usepackage[utf8]{inputenc}
\usepackage{csquotes}

\usepackage{microtype}
\usepackage{makecell,multirow}
\usepackage{amsmath,amssymb,amsthm}

\usepackage{xcolor}
\usepackage{wrapfig}
\usepackage{caption}
\usepackage{subcaption}

\usepackage{inconsolata}
\usepackage{todonotes}

\usepackage{graphicx}
\usepackage{tikz}
\usepackage{pgfplots}
\usepackage{float}
\restylefloat{table}
\usepackage{booktabs}

\input{macro}

\input{math_commands}

\newcommand*{\ethz}{$\zeta$}
\newcommand*{\google}{$\gamma$}

\title{Autoregressive Structured Prediction with Language Models}

\author{Tianyu Liu\textsuperscript{\ethz} \qquad Yuchen Eleanor Jiang\textsuperscript{\ethz} \\ \bf{Nicholas Monath}\textsuperscript{\google} \qquad \bf{Ryan Cotterell}\textsuperscript{\ethz} \qquad \bf{Mrinmaya Sachan}\textsuperscript{\ethz} \\

\setlength{\fboxsep}{2.pt}%
\setlength{\fboxrule}{2.pt}%

\fcolorbox{white}{white}{
\textsuperscript{\ethz}ETH Zürich \qquad \textsuperscript{\google}Google Research
} \\

\fcolorbox{white}{white}{
  $\{$\texttt{\href{mailto:tianyu.liu@inf.ethz.ch}{tianyu.liu},} \texttt{\href{mailto:yuchen.jiang@inf.ethz.ch}{yuchen.jiang}}$\}$\texttt{@inf.ethz.ch}
} \\
\fcolorbox{white}{white}{
\texttt{\href{mailto:nmonath@google.com}{nmonath@google.com}} \qquad $\{$\texttt{\href{mailto:ryan.cotterell@inf.ethz.ch}{ryan.cotterell},} \texttt{\href{mailto:mrinmaya.sachan@inf.ethz.ch}{mrinmaya.sachan}}$\}$\texttt{@inf.ethz.ch}
}
}

\begin{document}
\maketitle

\begin{abstract}
    In recent years, NLP has moved towards the application of language models to a more diverse set of tasks.
    However, applying language models to structured prediction, e.g., predicting parse trees, taggings, and coreference chains, is not straightforward.
    Prior work on language model-based structured prediction typically flattens the target structure into a string to easily fit it into the language modeling framework.
    Such flattening limits the accessibility of structural information and can lead to inferior performance compared to approaches that overtly model the structure.
    In this work, we propose to construct a conditional language model over sequences of structure-building actions, rather than over strings in a way that makes it easier for the model to pick up on intra-structure dependencies.
    Our method sets the new state of the art on named entity recognition, end-to-end relation extraction, and coreference resolution.\looseness=-1

    \vspace{1.5em}
    {\includegraphics[width=1.25em,height=1.25em]{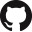}\hspace{1.5em}\parbox{\dimexpr\linewidth-2\fboxsep-2\fboxrule}{\url{https://github.com/lyutyuh/ASP}}}
    \vspace{-1.0em}
\end{abstract}

\input{sections/intro}

\input{sections/method}

\section{Experiments}
We experiment on three NLP structured prediction tasks: named entity recognition, end-to-end relation extraction, and coreference resolution.
We are primarily interested in understanding whether ASP provides advantages over two existing formalisms: (i) conditional language models \citep{athiwaratkun-etal-2020-augmented, tanl} that flatten the structure into a string (augmented language models), and (ii) the classic discriminative models whose autoregressivity is bounded.
We experiment with three pre-trained language models, T5 \cite{raffel-t5}, T0 \cite{sanh2021multitask}, and Flan-T5 \cite{flant5} for the three tasks under consideration.
Additional experimental details are given in \cref{appendix:exp_setting} and \cref{appendix:datasets}.\looseness=-1

\subsection{Named Entity Recognition}
First, we evaluate our model on the CoNLL-03 English NER task. Following previous work, we report the micro precision, recall, and F1 score. 
As shown in \cref{tab:result_test_ner_conll03}, our model using T0-3B backbone outperforms all other models without data augmentation or ensembling.

\input{tabs/result_test_ner}

\subsection{End-to-End Relation Extraction} 
We compare \asp{} on the CoNLL-04 and ACE-05 English end-to-end relation extraction datasets.
The results are shown in \cref{tab:result_test_ere_conll04} and \cref{tab:result_test_ere_ace05}.
Our proposed approach achieves state-of-the-art results on both datasets using T5-3B as the backbone.
In particular, it outperforms the flattened-string model of \citet{tanl} by a large margin ($>0.9$ F1).
We hypothesize that this is due to relations requiring higher-order dependencies between spans.

\input{tabs/result_test_ere}

\subsection{Coreference Resolution}
We then conduct experiments on the standard OntoNotes benchmark in the CoNLL-12 English shared task dataset \cite{pradhan-etal-2012-conll}.
\Cref{tab:result_test_coref} reports the results. Again, our model achieves state-of-the-art performance among systems without any data augmentation\footnote{We achieve $82.9$ F1 score on the development set, outperforming the result without pretraining on additional data reported by \citet[Tab. 5]{wu-etal-2020-corefqa}. 
In addition, training our model does not require the usage of TPUs.}, outperforming the previous state of the art by 1.5 F1 score.
We also observe that our \asp{} models substantially outperform discriminative models that make use of the same PLM.
Further analysis is provided in \cref{appendix:mention_recall}.

\input{tabs/result_test_coref}

\input{sections/related_work}

\section{Conclusion}
In this paper, we propose a novel framework for structured prediction
that encodes a structure as a series of structure-building actions that obtains state-of-the-art performance across three tasks.
In contrast to past approaches for structured prediction, our approach is compatible with pre-trained large language models.
This allows us to reduce structured prediction to the problem of fine-tuning pre-trained language models over an enlarged alphabet.
We show empirically that \asp{} outperforms previous structured prediction models by a large margin. 
Indeed, we set the new state of the art on three tasks: named entity recognition, end-to-end relation extraction, and coreference resolution.

\section*{Acknowledgements}
We acknowledge support from an ETH Z\"urich Research grant (ETH-19 21-1) and a grant from the Swiss National Science Foundation (project \#201009) for this work. We also thank Zeerak Talat and Peter Szemraj for their feedback on this manuscript. 

\section*{Ethical Considerations}
To consider the ethical implications of our work,
we consider the tasks and models used
and our proposed approach. The tasks considered, 
named entity recognition, relation extraction, and coreference resolution
are often used in a pipeline of approaches (say for automatically
building knowledge bases). Understanding the biases, errors, and failure cases of these tasks and their models and how they affect downstream use cases of the knowledge base would be important to consider. That said, to our knowledge the proposed approach does not exacerbate (or lessen) or introduce new considerations to the ones known about tasks/models more generally.

\section*{Limitations}

\paragraph{Autoregressive Modeling Assumption.}
The decoder model, which is autoregressive, introduces an inductive bias on the structured prediction approach.
Specifically, the left-to-right approach requires the model to model dependencies in a specific order.
This could account for some of the reduction in performance compared to task-specific discriminative models. Understanding the implications of the autoregressive decision is indeed an interesting question, but one that we felt was out of scope for this short paper.

\paragraph{Efficiency.}
In our experiments, we reduce the burden of finding many mention spans in two-stage approaches.  
On sentence-level tasks, e.g., entity and relation extraction, the number of decoding steps is relatively small. 
For instance, the average number of words in an input sentence is $\approx$20. Our system has a lighter memory trace as opposed to discriminative models. This extra time cost can be partially compensated with larger batch sizes.
However, on document-level tasks, e.g., coreference resolution, the number of decoding steps is too large to be compensated with parallelism. More efficient methods for inference such as non-autoregressive decoding \cite{gu2018nonautoregressive} remain to be explored in future work.

\paragraph{Decoding Algorithms.}
In this work, we use greedy decoding in all the experiments. 
Alternative decoding algorithms might further improve the quality of the generated sequences, e.g., beam search \cite{zhang-clark-2008-tale,goldberg-etal-2013-efficient}.

\paragraph{Choice of Pretrained Language Models.} 
In this work, the choice of T5 and its variants as the conditional language model backbone of our model is largely motivated by their ability to handle arbitrarily long sequences. Unlike BART and GPT, T5 uses relative position encoding. On document-level tasks such as coreference resolution, the ability to process long sequences is extremely important.
However, other pretrained conditional language models, either with encoder--decoder structures or decoder-only structures, can be used as a backbone. It might be interesting to explore techniques that generalize fixed-length position encoding to longer sequences.

\bibliography{anthology,custom}
\bibliographystyle{acl_natbib}
\clearpage
\appendix

\section{Experimental Details}
\subsection{Experimental Settings} \label{appendix:exp_setting}
In our experiments, the \{T5,T0,flan-T5\}-base, \{T5,T0,flan-T5\}-large, \{T5,T0,flan-T5\}-\{3B,XL\}, \{T5,T0,flan-T5\}-\{11B,XXL\} have 220 million, 770 million, 3 billion, and 11 billion parameters respectively\footnote{\url{https://github.com/google-research/text-to-text-transfer-transformer}}. The feedforward neural networks described in \cref{sec:instantiation} have one hidden layer of size 150 for ACE-05, 4096 for CoNLL-03, CoNLL-04, and CoNLL-12.\looseness=-1

We follow the same preprocessing procedure and train/dev/test split of previous work on all datasets. For all the experiments, we use the AdamW optimizer \cite{kingma2015adam}. 
We train 40 epochs on CoNLL-12 for coreference resolution with batch size 1. For end-to-end relation extraction on CoNLL-04 and ACE-05, we train 100 epochs with batch size 8. The initial learning rates are set to 5e-5 for \{T5,T0,flan-T5\}-base and \{T5,T0,flan-T5\}-large models, 3e-5 for \{T5,T0,flan-T5\}-\{3B,XL,11B,XXL\} models.\looseness=-1

We apply bfloat16 training in our experiments. One single A100-40GB GPU is used for training models that use \{T5,T0,flan-T5\}-base and \{T5,T0,flan-T5\}-large. Two A100-40GB GPUs are required to train models that use 3B or XL. Six A100-80GB GPUs are required to train models that use 11B or XXL models.
It takes around 0.1 seconds for base-scale models and 1 second per updating step for \{3B,11B,XXL\} models.\looseness=-1

\subsection{Datasets} \label{appendix:datasets}

\subsubsection{Named Entity Recognition}
\paragraph{CoNLL-03.}
We use the CoNLL-03 dataset \cite{tjong-kim-sang-de-meulder-2003-introduction} to evaluate our model on named entity recognition. This dataset consists of 946 training articles, 216 development articles, and 231 test sentences. We evaluate under the document-level settings, which means we feed the entire document into the model instead of the individual sentences.

\subsubsection{End-to-End Relation Extraction}
\paragraph{CoNLL-04.}
The CoNLL-04 dataset contains four types of entities (location, organization, person, other) and five types of relations (work for, kill, organization based in, live in, located in).
We split the dataset as the training (922 sentences), validation (231 sentences), and test (288 sentences) as in previous work.
For the ACE-05 dataset, we follow the train/dev/test split of previous work \cite{zhong-chen-2021-frustratingly}.

\paragraph{ACE-05.}
The ACE-05 dataset \cite{walker2005ace} contains 511 documents in total collected from multiple domains including newswire, broadcast, discussion forums, etc. We follow \citet{luan-etal-2019-general}'s preprocessing script\footnote{https://github.com/luanyi/DyGIE/tree/master/preprocessing} and split the dataset into train/dev/test set.
ACE-05 contains inconsistently capitalized data. The newswire portion collected from CNN are entirely lowercased, which involves around 20 documents. Previous works \cite{zhong-chen-2021-frustratingly,ye-etal-2022-packed} that use case-insensitive encoders such as ALBERT are not affected by this deficiency. 
However, the T5 model and its variants are case-sensitive. We use the python \texttt{truecase} package\footnote{https://pypi.org/project/truecase/} to restore the correct capitalization during preprocessing.

\subsubsection{Coreference Resolution}
\paragraph{CoNLL-12.}
The CoNLL-12 English shared task dataset for coreference resolution \cite{pradhan-etal-2012-conll} contains 2802 documents for training, 343 for validation, and 348 for testing. 
During training, we chunk the documents into segments of 2048 maximum words. 
In total, there are 2830 segments for training.
During the evaluation, we use the entire document as the input to the model.

\begin{figure}[h!]
\begin{center}
\begin{tikzpicture}[scale=0.9]
    \begin{axis}[
        xlabel= Ratio,
        ylabel= Recall (\%),
        y label style={at={(axis description cs:0.05,0.5)}},
        x label style={at={(axis description cs:0.5,0.03)}},
        xmajorgrids=true,
        ymajorgrids=true,
        legend cell align={left},
        legend style={at={(1,0.3)}},
        height=7cm
    ]
      \addplot [mark=none,color=blue] 
      coordinates {
        (0.1,  65.91)
        (0.2,  90.36)
        (0.25, 93.74)
        (0.3,  95.04)
        (0.4,  96.24)
        (0.5,  96.75)
        (0.6,  97.15)
    };
    \addplot [mark=*,color=blue,only marks] 
      coordinates {
        (0.4,  96.24)
    };
    \addplot [mark=square*,color=red,only marks] 
      coordinates {
        (0.0959,  89.43)
    };
    \legend{\texttt{Joshi} (various ratio)\\ \texttt{Joshi} (actual ratio)\\ $\texttt{ASP}+\tfive_\textsc{l}$\\}
    \end{axis}
\end{tikzpicture} 
\end{center}
\caption{Recall rate of gold mentions. 
The ratio on the $x$-axis refers to the number of predicted mentions divided by $|D|$. \texttt{Joshi} refers to the two-stage model of \citep{joshi-etal-2020-spanbert}.}
\label{fig:mention_detection}
\end{figure}

\input{tabs/result_test_conll}

\section{Coreference Resolution} \label{appendix:mention_recall}
In this section, we analyze the performance of mention detection for coreference resolution of our model in \cref{fig:mention_detection}. This analysis casts light on how our model \emph{plans globally} in an autoregressive manner.
In the task of coreference resolution, only the entities that are mentioned more than once in a document are annotated as mentions. This is to say, an utterance of an entity should only be labeled if that entity is referred to again afterward. Thus, in previous coreference resolution models, a dedicated mention detection module that enumerates candidate textual spans (e.g., noun phrases and pronouns) for mentions is indispensable. 
However, our model is able to directly predict the exact set of mentions that we require, even if the target sequence is generated from left to right. We conclude that this results from the cross-attention mechanism which enables the model to look at relevant parts in the input document during decoding.
Given an input document of $|D|$ words, our model predicts only $0.096\,|D|$ mentions with a $89.6\%$ recall rate of gold mentions. This refrained mention detection strategy imposes a limit on the cardinality of $\calZ_i$ in \cref{eq:support_general}. As a result, this relatively small constant factor (compared to 0.4 used in most previous work) keeps our model tractable without the need for pruning strategies as in the models based on \cite{lee-etal-2017-end}.

\section{Modeling More Restricted Structures} \label{appendix:parsing}
In this work, we tackled three tasks that are traditionally considered structured prediction problems. Named entity recognition and relation extraction consider labeling spans with a set of given types. 
Coreference resolution has long-range dependencies and has to model relationships between spans. 
However, there are structured prediction problems that require more restricted outputs. For instance, in dependency parsing, a spanning tree connecting every word in the input sentence is the desired output \cite{dependencyparsing}. While in constituency parsing, a parse tree in Chomsky Normal Form is supposed to be a complete binary tree except for the leaf nodes \cite{handbook-clnlp}.
Modeling such types of structures requires a more specified definition of task-specific actions. In future work, we aim to explore the abilities and limitations of our method.\looseness=-1

\section{Experiments with Flan-T5}
We conduct additional experiments with the latest pretrained language model Flan-T5 \cite{flant5}. Flan-T5 is pretrained on more supervised tasks and achieves better performance than the original T5 on multiple NLU tasks. 
The results are shown in \cref{tab:result_test_conll}, \cref{tab:result_test_ner_conll03_flant5}, and \cref{tab:result_test_ere_conll04_flant5}. We find that with the same size of the model, Flan-T5 performs better than T5 in general.

\begin{table}[t]
\centering
\resizebox{0.45\textwidth}{!}{
\small
\begin{tabular}{lccc}
\toprule
             & Prec. & Rec. & F1 \\ \midrule
\texttt{ASP}$+\tfive_\textsc{b}$ & 91.4 & 92.2 & 91.8 \\ 
{\texttt{ASP}}$+{\tt{\flant_\textsc{b}}}$ & 92.7 & 93.8 & 93.3 \\ 
{\texttt{ASP}}$+\tfive_\textsc{l}$ & 92.1  & 93.4  & 92.8 \\
{\texttt{ASP}}$+{\tt{\flant_\textsc{l}}}$ & 93.3 & 94.3 & 93.8 \\  
{\texttt{ASP}}$+\tfive_\textsc{3b}$ & {93.8}  & {94.4} &  \textbf{94.1} \\ \bottomrule 
\end{tabular}} 
\caption{Test F1 scores of named entity recognition on the CoNLL-03 test set.}
\label{tab:result_test_ner_conll03_flant5}
\end{table}

\begin{table}[t]
\centering

\begin{tabular}{lcc}
\toprule
             & Ent & Rel \\ \midrule
\texttt{ASP}$+\tfive_\textsc{b}$ & 89.5 & 73.2\\ 
{\texttt{ASP}}$+{\tt{\flant_\textsc{b}}}$ & 89.4 & 73.8\\ 
{\texttt{ASP}}$+{\tt{\flant_\textsc{l}}}$ & 90.5 & 76.2 \\ \bottomrule
\end{tabular}
\caption{Test F1 scores of named entity recognition on the CoNLL-04 test set.}
\label{tab:result_test_ere_conll04_flant5}
\end{table}

\begin{table*}[h]
    \small
    \centering 
\resizebox{\textwidth}{!}{
    \begin{tabular}{p{1.6cm} p{\textwidth-1.6cm}} \toprule
       \textbf{named entity recognition}  &  \texttt{GUNMEN WOUND TWO $\actionlsb{}$ MANCHESTER UNITED $\actionrsb{}$ FANS IN $\actionlsb{}$ AUSTRIA $\actionrsb{}$. $\actionlsb{}$ VIENNA $\actionrsb{}$ 1996-12-06 Two $\actionlsb{}$ Manchester United $\actionrsb{}$ soccer fans were wounded by unidentified gunmen on Friday and taken to hospital in the $\actionlsb{}$ Austrian $\actionrsb{}$ capital, police said. " The four $\actionlsb{}$ Britons $\actionrsb{}$ were shot at from a $\actionlsb{}$ Mercedes $\actionrsb{}$ car at around 1 a.m., " a spokeswoman told $\actionlsb{}$ Reuters $\actionrsb{}$. The two men were hit in the pelvis and leg. Police said their lives were not in danger. The fans, in $\actionlsb{}$ Austria $\actionrsb{}$ to watch their team play $\actionlsb{}$ Rapid Vienna $\actionrsb{}$ last Wednesday, may have been involved in a pub brawl earlier, the spokeswoman said. $\actionlsb{}$ Manchester United $\actionrsb{}$ won 2-0.</s>} \\ \midrule
       \textbf{end-to-end relation extraction}  &  \texttt{And this final story: retired $\actionlsb{}$ Senator $\actionrsb{}$ $\actionlsb{}$ Strom Thurmond $\actionrsb{}$ has never made a secret about $\actionlsb{}$ his $\actionrsb{}$ fondness for young pretty $\actionlsb{}$ women $\actionrsb{}$ .</s>} \\ \midrule
       \textbf{coreference resolution}  &  \texttt{<speaker> - </speaker> $\actionlsb{}$ Al Gore $\actionrsb{}$ won't be the next U.S. President, but $\actionlsb{}$ he $\actionrsb{}$  has a slim chance of becoming $\actionlsb{}$ the next President at $\actionlsb{}$ Harvard $\actionrsb{}$ $\actionrsb{}$. $\actionlsb{}$ Gore $\actionrsb{}$ holds a degree from $\actionlsb{}$ the university $\actionrsb{}$, and is one of about 500 people nominated for $\actionlsb{}$ the job $\actionrsb{}$. $\actionlsb{}$ A school official $\actionrsb{}$ talked about $\actionlsb{}$ the Vice President's $\actionrsb{}$ chances during an interview with " the Boston Globe. "  $\actionlsb{}$ He $\actionrsb{}$ says it's unlikely $\actionlsb{}$ Gore $\actionrsb{}$ will be selected,  because $\actionlsb{}$ he $\actionrsb{}$ doesn't have enough experience in the academic world.</s>} \\ \cmidrule{2-2}
       & \texttt{<speaker> - </speaker> $\actionlsb{}$ Violence between Israelis and Palestinians $\actionrsb{}$ continued in $\actionlsb{}$ its $\actionrsb{}$ third month, though at a slightly reduced level overall. $\actionlsb{}$ Israeli and Palestinian negotiators $\actionrsb{}$ met separately at the White House with President Bill Clinton in hopes of restarting direct negotiations between $\actionlsb{}$ them $\actionrsb{}$ for a final settlement.</s>} \\
       \bottomrule
    \end{tabular}
    }
    \caption{Predicted sequences from CoNLL-03, ACE-05, and CoNLL-12 dataset.}
    \label{tab:qualitative_example}
\end{table*}

\section{Decoding Examples}
We provide decoding examples from the tasks we experiment on in \cref{tab:qualitative_example}. The $\actioncopy$ actions are verbalized into tokens.

\end{document}

%% file: macro.tex
\newcommand{\bigO}[1]{\mathcal{O}(#1)}

\newcommand*{\defeq}{\mathrel{\stackrel{\textnormal{\tiny def}}{=}}}
\newcommand{\defn}[1]{\textbf{#1}}

\newcommand*{\asp}{\texttt{ASP}}
\newcommand*{\tfive}{\textsc{t{\scriptsize 5}}}
\newcommand*{\tzero}{\textsc{t{\scriptsize 0}}}
\newcommand*{\flant}{\textsc{flan}\text{-}\textsc{t{\scriptsize 5}}}

\definecolor{mygreen}{HTML}{7FC07F}
\definecolor{myred}{HTML}{DF7F7F}
\definecolor{mygray}{HTML}{BAB7B7}
\definecolor{pine}{HTML}{BCE672}
\definecolor{lotus}{HTML}{E4C6D0}
\definecolor{gold}{HTML}{F2BE45}
\definecolor{bamboo}{HTML}{789262}
\definecolor{jade}{HTML}{3DE1AD}
\definecolor{blueblue}{HTML}{4C8DAE}

\newcommand*{\actioncopy}{\texttt{{\setlength{\fboxsep}{2.5pt}\colorbox{mygray}{copy}}}}
\newcommand*{\actionlsb}{\texttt{{\setlength{\fboxsep}{2pt}\colorbox{mygreen}{[$\hspace{-1pt}^*$}}}}
\newcommand*{\actionrsb}{\texttt{{\setlength{\fboxsep}{2pt}\colorbox{myred}{]}}}}

\newcommand*{\calA}{\mathcal{A}}
\newcommand*{\calB}{\mathcal{B}}
\newcommand*{\calC}{\mathcal{C}}
\newcommand*{\calE}{\mathcal{E}}

\newcommand*{\calZ}{\mathcal{Z}}
\newcommand*{\calY}{\mathcal{Y}}
\newcommand*{\calYn}{\mathcal{Y}_n}
\newcommand*{\calL}{\mathcal{L}}

\newcommand*{\calR}{\mathcal{R}}
\newcommand*{\calT}{\mathcal{T}}

\newcommand*{\ptheta}{p_{\boldsymbol{\theta}}}
\newcommand*{\stheta}{s_{\boldsymbol{\theta}}}

\usepackage{cleveref}
\crefname{section}{\S}{\S\S}
\Crefname{section}{\S}{\S\S}
\crefname{table}{Tab.}{}
\crefname{figure}{Fig.}{Figs.}
\crefname{algorithm}{Algorithm}{}
\crefname{equation}{Eq.}{Eqs.}
\crefname{line}{Line}{}
\crefname{appendix}{App.}{}
\crefformat{section}{\S#2#1#3}
\crefname{thm}{Theorem}{}
\crefname{cor}{Corollary}{}
\crefname{prop}{Proposition}{}
\crefname{def}{Definition}{}

\makeatletter
\newcommand*\iftodonotes{\if@todonotes@disabled\expandafter\@secondoftwo\else\expandafter\@firstoftwo\fi}
\makeatother


%% file: math_commands.tex
\usepackage{amsmath,amsfonts,bm}

\def\eqref#1{equation~\ref{#1}}

\def\1{\bm{1}}

\def\ry{{\boldsymbol{y}}}
\def\rz{{\textnormal{z}}}

\def\rvy{{\boldsymbol{y}}}

\def\vh{{\bm{h}}}

\def\vm{{\bm{m}}}

\def\vw{{\bm{w}}}

\DeclareMathAlphabet{\mathsfit}{\encodingdefault}{\sfdefault}{m}{sl}
\SetMathAlphabet{\mathsfit}{bold}{\encodingdefault}{\sfdefault}{bx}{n}

\DeclareMathOperator*{\argmax}{argmax}

%% file: sections/intro.tex
\begin{figure*}[!ht]
     \centering
     \includegraphics[width=\textwidth,trim={0.3cm 0.cm 0.5cm 0.1cm} ,clip]{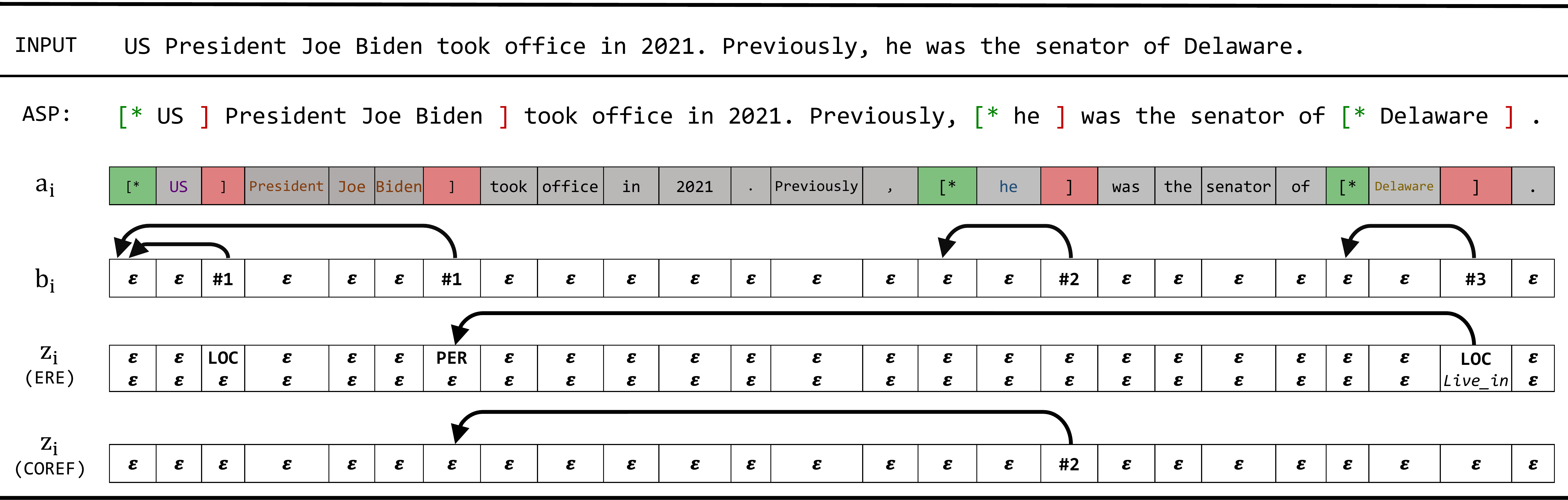}
        \caption{Illustration of the target outputs of our framework on coreference resolution (\textbf{\textsc{coref}}) and end-to-end relation extraction (\textbf{\textsc{ere}}). The lower part illustrates the decoding process of our model. The actions $\ry_i$ are color-coded as \actionrsb{}, \actionlsb{} and \actioncopy{}. The structure random variables $\rz_i$ are presented along with coreference links or relation links. We present words in the \actioncopy{} cells merely as an illustration. \looseness=-1 } 
        \label{fig:illustration}
\end{figure*}

\section{Introduction}
Many common NLP tasks, e.g., named entity recognition, relation extraction, and coreference resolution are naturally taxonomized as structured prediction, the supervised machine-learning task of predicting
a structure from a large\footnote{Typically, large means exponential in the size of the input.} set.
To generalize well to held-out data in a structured prediction problem, the received wisdom has been that it is necessary to correctly model complex dependencies between different pieces of the structure.
However, a recent trend in structured prediction for language has been to forgo explicitly modeling such dependencies \citep[\textit{inter alia}]{ma-hovy-2016-end, lee-etal-2017-end, he-etal-2017-deep}, and, instead, to apply an expressive black-box model, e.g., a neural network, with the hope that the model picks up on the dependencies without explicit instruction.\looseness=-1

Framing structured prediction as conditional language modeling is an increasingly common black-box technique for building structured predictors that has led to empirical success \citep[\emph{inter alia}]{NIPS2015_277281aa, raffel-t5,athiwaratkun-etal-2020-augmented,decao2020autoregressive,tanl}. 
The idea behind the framework is to encode the target structure as a string, flattening out the structure.
Then, one uses a conditional language model to predict the flattened string encoding the structure.
For instance, \citet{NIPS2015_277281aa} flatten parse trees into strings and predict the strings encoding the flattened trees from the sentence with a machine translation architecture.
The hope is that the autoregressive nature of the language model allows it to \emph{learn} to model the intra-structure dependencies and the necessary hard constraints that ensure the model even produces well-formed structures.
Additionally, many modelers make use of pre-trained language models \cite{lewis,raffel-t5} to further improve the language models.\looseness=-1

However, despite their empirical success, simply hoping that a black-box approach correctly models intricate intra-structure dependencies is often insufficient for highly structured tasks \citep[\S1]{tanl}.
Indeed, the act of flattening a structured object into a string
makes properly modeling the intra-structure dependencies harder for many tasks, e.g., those that involve nested spans or long-distance dependencies. 
For instance, in coreference resolution, a conference link between two mentions can stretch across thousands of words, and a coreference chain can also contain over a hundred mentions \citep{pradhan-etal-2012-conll}.
Flattening such a large amount of structured information into a string makes the task more difficult to model.\looseness=-1

In this paper, we propose a simple framework that augments a conditional language model with explicit modeling of structure.
Instead of modeling strings that encode a flattened representation of the target structure, we model a constrained set of actions that build the target structure step by step; see \cref{fig:illustration} for an example of our proposed framework.
Training a conditional language model to predict structure-building actions exposes the structure in a way that allows the model to pick up on the intra-structure dependencies more easily while still allowing the modeler to leverage pre-trained language models.
We conduct experiments on three structured prediction tasks: named entity recognition, end-to-end relation extraction, and coreference resolution. 
On each task, we achieve state-of-the-art results \emph{without} relying on data augmentation or task-specific feature engineering.\looseness=-1

%% file: sections/method.tex
\section{Autoregressive Structured Prediction} \label{sec:model}

In this section, we describe our proposed approach, which we refer to as \textbf{autoregressive structured prediction} (\asp).
Unlike previous approaches for structured prediction based on conditional language modeling, we represent structures as sequences of \defn{actions}, which build pieces of the target structure step by step.
For instance, in the task of coreference resolution, the actions build spans as well as the relations between the spans, contiguous sequences of tokens.
We give an example in \cref{fig:illustration}.

\subsection{Representing Structures with Actions}
 While our approach to structured prediction, \asp, is quite general, our paper narrowly focuses on modeling structures that are expressible as a set of dependent spans, and we couch the technical exposition in terms of modeling spans and relationships among spans.
Our goal is to predict an action sequence $\rvy=y_1, \ldots, y_{N}$, where each action $y_n$ is chosen from an \defn{action space} $\calYn$.
In this work, we take $\calYn$ to be factored, i.e., $\calYn \defeq \calA \times \calB_n \times \calZ_n$, where $\calA$ is a set of structure-building actions, $\calB_n$ is the set of bracket-pairing actions, and $\calZ_n$ is a set of span-labeling actions. 
Thus, each $y_n$ may be expressed as a triple, i.e., $y_n = \langle a_n, b_n, z_n \rangle$.
We discuss each set in a separate paragraph below.\looseness=-1

\paragraph{Structure-Building Actions.} 
We first define a set of structure-building actions $\calA = \Big\{\actionrsb, \actionlsb, \actioncopy\Big\}$ that allow us to encode the span structure of a text, e.g., $\actionlsb{}\textit{Delaware}\actionrsb{}$ in \cref{fig:illustration} encodes that \textit{Delaware} is a span of interest.
More technically, the action $\actionrsb$ refers to a right bracket that marks the right-most part of a span.
The action $\actionlsb$ refers to a left bracket that marks the left-most part of a span.
The superscript ${}^*$ on $\actionlsb$ is inspired by the Kleene star and indicates that it is a placeholder for 0 or more consecutive left brackets\footnote{In our preliminary experiments, we observe unsatisfactory performance when the model has to generate consecutive left brackets. We leverage $\actionlsb{}$ as an engineering workaround. We hypothesize that this phenomenon is due to the inability of transformers to recognize \texttt{Dyck} languages~\citep{hahn-2020-theoretical, hao-etal-2022-formal}.}.
Finally, $\actioncopy$ refers to copying a word from the input document. 
To see how these actions come together to form a span, consider the subsequence in \cref{fig:illustration}, $\actionlsb{}\textit{Delaware}\actionrsb{}$, which is generated from a sequence of structure-building actions $\actionlsb{}$, $\actioncopy{}$, and $\actionrsb{}$.\looseness=-1

\paragraph{Bracket-Pairing Actions.}
Next, we develop the set of actions that allow the model to match left and right brackets; we term these bracket-pairing actions. 
The set of bracket-pairing actions consists of
all previously constructed left brackets, i.e.,
\begin{align}
    \calB_n = \Big\{m \mid m < n \land a_m = \actionlsb{}\Big\}
\end{align}
Thus, in general, $|\calB_n|$ is $\bigO{n}$.
However, it is often the case that domain-specific knowledge can be used to prune $\calB_n$.
For instance, coreference mentions and named entities rarely cross sentence boundaries, which yields a linguistically motivated pruning strategy \citep{liu-etal-2022-structured}.
Thus, in some cases, the cardinality of $\calB_n$ can be significantly smaller.
When we decode action sequences $\ry$ into a structure, unpaired $\actionlsb{}$ and $\actionrsb{}$ can be removed ensuring that the output of the model will not contain unpaired brackets.\looseness=-1

\paragraph{Span-Labeling Actions.}
Finally, we add additional symbols $z_n$ associated with each $y_n$ that encode a labeling of a single span or a relationship between two or more spans.
For instance, see \Cref{sec:instantiation} for an example.
We denote the set of all $z_n$ as
\begin{align} \label{eq:support_general}
    \calZ_n = \Big\{m \mid m < n \land a_m = \actionrsb{} \Big\} \times \calL
\end{align}
where $\Big\{m \mid m < n \land a_m = \actionrsb{} \Big\}$ is the set of previous spans, which allows the model to capture intra-span relationships, and $\calL$ denotes the set of possible labelings of the current span and the relationship between the adjoined spans. 
In general, designing $\calZ_n$ requires some task-specific knowledge in order to specify the label space.
However, we contend it requires less effort than designing a flattened string output where different levels of structures may be intertwined \cite{tanl}. \looseness=-1

\subsection{Model Parameterization}
Let $D = \vw_1, \dots, \vw_{K}$ be an input document of $K$ sentences where $\vw_k$ denotes the $k^{\text{th}}$ sentence in $D$.
We first convert the structure to be built on top of $D$ into an action sequence, which we denote as $\ry$ where $y_n \in \calY_n$. 
Now, we model the sequence of actions $\ry$ as a conditional language model\looseness=-1
\begin{equation}
    \ptheta(\rvy \mid D) = \prod_{n=1}^N \ptheta(y_n \mid \rvy_{< n}, D)
\end{equation}
The log-likelihood of the model is then given by $\log \ptheta(\rvy \mid D) = \sum_{n=1}^N \log \ptheta(y_n \mid \rvy_{<n}, D)$.
We model the local conditional probabilities $p(y_n \mid \rvy_{<n}, D)$ as a softmax over
a \emph{dynamic} set $\calYn$ that changes as a function of the history $\rvy_{<n}$, i.e.,
\begin{align}\label{eq:loglinear}
    \ptheta(y_n \mid \rvy_{<n}, D) = \frac{\exp \stheta(y_n)}{\sum_{y_n^{\prime} \in \calY_n} \exp \stheta(y_n^\prime)}
\end{align}
where $\stheta$ is a parameterized score function; we discuss several specific instantiations of $\stheta$ in \cref{sec:instantiation}.
Finally, we note that the use of a dynamic vocabulary stands in contrast to most conditional language models where the vocabulary is held constant across time steps, e.g., \citeposs{sutskever2014sequence} approach to machine translation.\looseness=-1

\paragraph{Greedy Decoding.}
We determine the approximate best sequence $\rvy^*$ using a greedy decoding strategy.
At decoding step $n$, we compute
\begin{equation}
    y_n^* = \argmax_{y'_n} \ptheta(y'_n \mid \rvy_{<n}, D)
\end{equation}
The chosen $y_n^* = \langle a_n^*, b_n^*, z_n^*\rangle$ will then be verbalized as a token as follows: If $a_n^* = \actioncopy$, then we copy the next token from the input that is not present in the output. 
Otherwise, if $y_n^* = \actionlsb$ or $y_n^* = \actionrsb$, we insert $\actionlsb$ or $\actionrsb$ into the output sequence, respectively. 
The verbalized token is then fed into the conditional language model at the next step. 
The decoding process terminates when the model copies a distinguished symbol \textsc{eos} symbol from the input.
The end of the procedure yields an approximate argmax $\rvy^*$.

\looseness=-1

\paragraph{Computational Complexity.}
\cref{eq:loglinear} can be computed quite efficiently using our framework, as the cardinalities of $\calA$ is $\bigO{1}$, and the size of $\mathcal{B}_n$ and $\mathcal{Z}_n$ are both $\bigO{n}$.
A tighter analysis says the cardinalities of $\mathcal{B}_n$ and $\mathcal{Z}_n$ are roughly linear in the number of spans predicted. 
In practice, we have $n \ll |V|$ where $|V|$ is the size of vocabulary, which is the step-wise complexity of \cite{tanl}.
A quantitative analysis of the number of mentions in coreference can be found in \cref{appendix:mention_recall}. \looseness=-1

\paragraph{Generality.}
Despite our exposition's focus on tasks that involve assigning labels to span or span pairs, our method is quite general. 
Indeed, almost any structured prediction
task can be encoded by a series of structure-building actions.
For tasks that involve labeling tuples of spans, e.g., semantic role labeling makes use of tree-tuples that consist of the subject, predicate, and object, \cref{eq:support_general} can be easily extended with a new space of categorical variables $\Big\{ m \mid m < n \land a_m = \actionrsb{} \Big\}$ to model the extra item.

\subsection{Task-specific Parameterizations} \label{sec:instantiation}
We now demonstrate how to apply \asp{} to three language structured prediction tasks: 
named entity recognition, coreference resolution, and end-to-end relation extraction.

\paragraph{Named Entity Recognition.} 
Named entity recognition is the task of labeling all mention spans  $\calE = \{ e_n \}_{n=1}^{|\calE|}$ in a document $D$ that refers to named entities. 
Since named entity recognition only requires labeling spans (and not linking them), we only need our task-specific $z_n$ to encode the entity type, which is canonically taken from a set of pre-defined categories $\calC$.
The function $\stheta(y_n)$ in \cref{eq:loglinear} is implemented by a feed-forward network
\begin{align}\label{eq:step_scorer_ner} 
    \stheta(y_n &= \langle a_n, b_n, z_n \rangle )  \\
    &\defeq 
    \begin{cases}
        \text{FFN}_{a_n}^{z_n}(\vm_n) & \textbf{if } a_n = \actionrsb \\
        \text{FFN}_{a_n}(\vh_n) & \textbf{otherwise}
    \end{cases} \nonumber
\end{align}
where $\vh_n$ is the decoder hidden state at step $n$, a column vector, and $\vm_{n}=[\vh_n^{\top}; \vh_{b_n}^{\top}]^{\top}$ represents the mention that corresponds to $y_n$.
Note that each $\text{FFN}_{a_n}^{z_n}$ and $\text{FFN}_{a_n}$ represent independent feed-forward networks with \emph{no} shared parameters.\looseness=-1

\paragraph{End-to-End Relation Extraction.} 
End-to-end relation extraction is the task of jointly extracting a set of entities alongside a set of relations between pairs of extracted entities.
Formally, given a set of pre-defined entity categories $\calC$ and a set of pre-defined relations $\calR$.
The goal is (i) to identify all possible entities $\calE = \{ e_n\}_{n=1}^{|\mathcal{E}|}$ in $D$ that could be associated with one of the entity types $c$ in $\calC$ and (ii) to identify all possible triples $\calT = \{ (e_n, r_n, e'_n) \}_{n=1}^{|\calT|}$ in $D$ where $e_n, e'_n \in \calE$ are the head and tail entity and $r_n \in \calR$ is the relation between $e_n$ and $e'_n$. 
Here, the support of $z_n$ takes the form of \cref{eq:support_general}, where $\calL$ is instantiated as $\calC \times \calR$. 
And $\stheta(y_n)$ kept the same as in \cref{eq:step_scorer_ner}.

\paragraph{Coreference Resolution.} 

The task of coreference resolution involves identifying all mention spans $\calE = \{ e_n \}_{n=1}^{|\calE|}$ in $D$ and then clustering them.
However, in addition to identifying the mention spans, the task of coreference resolution requires us to assign an antecedent to every possible mention in $D$.
To encode coreference resolution in our framework, we consider the task-specific $z_n$ from the set\looseness=-1
\begin{align}\label{eq:support_z_coref}
    \calZ_n =  \Big\{m \mid m < n \land a_m = \actionrsb \Big\} \cup \{\epsilon\}
\end{align}
where we follow the convention set in \citet{lee-etal-2017-end} that the antecedent of the first mention in each coreference chain is defined to be $\epsilon$.
Again, we define $\stheta(y_n = \langle a_n, b_n, z_n \rangle)$ as in \cref{eq:step_scorer_ner} with the exception that, when $z_n = \epsilon$, we define $\text{FFN}_{a_n}(\vm_n)_{\epsilon}=\text{FFN}_{a_n}(\vm_n)$. \looseness=-1

%% file: tabs/result_test_ner.tex
\begin{table}[t]
\centering
\resizebox{0.5\textwidth}{!}{
\small
\begin{tabular}{lccc}
\toprule
            & Prec. & Rec. & F1 \\ \midrule
 \citet{ma-hovy-2016-end} & 91.4 & 91.1  & 91.2 \\
  \citeauthor{devlin-etal-2019-bert}$+{\textsc{bert}_\textsc{l}}$ & - & - &  92.8 \\  \citeauthor{ye-etal-2022-packed}$+{\textsc{roberta}_\textsc{l}}$ & - & - & 94.0  \\ 
\citeauthor{athiwaratkun-etal-2020-augmented} & - & - & 91.5 \\ 
\citeauthor{tanl}$+{\tt{\tfive_\textsc{b}}}$  & - & - & 91.7 \\ \midrule
\texttt{ASP}$+\tfive_\textsc{b}$ & 91.4 & 92.2 & 91.8 \\ 
{\texttt{ASP}}$+\tfive_\textsc{l}$ & 92.1  & 93.4  & 92.8 \\
{\texttt{ASP}}$+\tfive_\textsc{3b}$ & {93.8}  & {94.4} &  \textbf{94.1} \\ \bottomrule
\end{tabular}} 
\caption{Test F1 scores of named entity recognition on the CoNLL-03  test set.}
\label{tab:result_test_ner_conll03}
\end{table}

%% file: tabs/result_test_ere.tex
\begin{table}[ht]
\centering
\resizebox{0.5\textwidth}{!}{
\small
\begin{tabular}{lcc}
\toprule
                 & Ent  & Rel  \\ \midrule
\citet{eberts2019span} & 88.9 & 71.5   \\ 
 \citet{ijcai2020-546} & 88.9 & 71.9    \\ 
 \citeauthor{wang-lu-2020-two}$+{\textsc{albert}_\textsc{xxl}}$ & 90.1 & 73.8    \\ 
\citeauthor{tanl}$+\tfive_\textsc{b}$ & 89.4 & 71.4  \\ \midrule
 \texttt{ASP}$+\tfive_\textsc{b}$& 89.5  & 73.2   \\ 
 \texttt{ASP}$+\tzero_\textsc{3b}$ & \textbf{90.3}  & \textbf{76.3}   \\ \bottomrule
\end{tabular}} 
\caption{\textbf{Micro} F1 scores of entity extraction and relation extraction on the CoNLL-04 joint entity relation extraction test set.}
\label{tab:result_test_ere_conll04}
\end{table}

\begin{table}[ht]
\centering
\resizebox{0.5\textwidth}{!}{
\small
\begin{tabular}{lccc}
\toprule
                   & Ent & Rel & Rel+ \\ \midrule
\citeauthor{wang-lu-2020-two}$+{\textsc{alb}_\textsc{xxl}}$ & 89.5 & 67.6  & 64.3 \\
\citeauthor{zhong-chen-2021-frustratingly}$+{\textsc{alb}_\textsc{xxl}}$ & 90.9 & 69.4 &  67.0 \\  
\citeauthor{ye-etal-2022-packed}$+{\textsc{alb}_\textsc{xxl}}$\footnotemark{} & 91.1 & 72.4 & 70.3  \\ 
\citeauthor{tanl}$+\tfive_\textsc{b}$ & 88.9 & 63.7 & - \\ \midrule
 \texttt{ASP}$+\tfive_\textsc{b}$ & 90.7  & 71.1  & 68.6 \\ 
 \texttt{ASP}$+\tfive_\textsc{l}$ & 91.3  & 71.9  & 69.4 \\ 
 \texttt{ASP}$+\tfive_\textsc{3b}$\footnotemark{} & \textbf{91.3}  & \textbf{72.7} &  \textbf{70.5} \\ \bottomrule
\end{tabular}} 
\caption{Test F1 scores of entity and relation extraction on the ACE-05 joint entity relation extraction task.}
\label{tab:result_test_ere_ace05}
\end{table}
\footnotetext[\numexpr\value{footnote}-1]{\citet{ye-etal-2022-packed} counts symmetric relations twice for evaluation, which is inconsistent with previous work. We report the re-evaluated scores under the standard metric.}
\footnotetext{On ACE-05, we observe inferior performance using T0-3B instead of T5-3B. We suspect this is due to systematic deficiencies in dataset preprocessing, e.g., errors during sentencization and tokenization as well as inconsistent capitalization.\looseness=-1}

%% file: tabs/result_test_coref.tex
\begin{table}[t]
\centering
\resizebox{0.5\textwidth}{!}{
\small
\begin{tabular}{lcccc}
\toprule
                     & MUC    & B$^3$   & CEAF$_{\phi_4}$  &      Avg. F1    \\ \midrule
\citet{lee-etal-2017-end} & 75.8 & 65.0  & 60.8          & 67.2          \\ 
 \citet{joshi-etal-2020-spanbert} & 85.3   & 78.1         & 75.3          & 79.6           \\
 \citeauthor{joshi-etal-2020-spanbert}$+\tfive_\textsc{b}$ $^\dagger$ \hspace{-0mm} &  79.8         & 70.2          & 66.8          & 72.3       \\ 
 \citeauthor{joshi-etal-2020-spanbert}$+\tfive_\textsc{l}$ $^\dagger$ & 81.4          & 73.1         & 73.1          & 74.9     \\

\citeauthor{urbizu-etal-2020-sequence} & 
64.9 &  66.5 & 65.3 & 65.6   \\ 
 \citeauthor{tanl}$+\tfive_\textsc{b}$  & 81.0 & 69.0 & 68.4 & 72.8   \\ 
 \citeauthor{dobrovolskii-2021-word} & 86.3          & 79.9         & 76.6          & 81.0     \\ \midrule
\texttt{ASP}$+\tfive_\textsc{b}$ & 82.3 & 75.1 & 72.5 &  76.6  \\ 
\texttt{ASP}$+\tfive_\textsc{l}$ & 84.7 & 77.7  & 75.2 & 79.3  \\ 
\texttt{ASP}$+\tzero_\textsc{3b}$ & 86.9  & 81.5  & 78.4 & 82.3 \\ 
\texttt{ASP}$+\flant_\textsc{xxl}$ & 87.2 & 81.7  & 78.6 & \textbf{82.5}\\ \bottomrule
\end{tabular}} 
\caption{Results on the CoNLL-12 English test set. Avg. F1 denotes the average F1 of MUC, B{$^3$}, and CEAF{$_{\phi_4}$}. 
Models marked with ${}^\dagger$ are our re-implementation. Other results are taken from their original papers. The full results are in \Cref{tab:result_test_conll}.}
\label{tab:result_test_coref}
\end{table}

%% file: sections/related_work.tex
\section{Related Work}

Most similar to our approach is the model of  \cite{tanl}, which also predicts structures in an iterative manner using conditional language models.
Similar approaches exist for constituency parsing \cite{NIPS2015_277281aa,dyer-etal-2016-recurrent}, entity retrieval \cite{decao2020autoregressive},
semantic parsing \cite{xiao2016sequence}, slot labeling, and intent classification \cite{athiwaratkun-etal-2020-augmented}. 
Earlier work on search-based \cite{daume2009search,doppa2014structured,chang2015learning} and greedy-based approaches \cite{swayamdipta2016greedy} applied to structured prediction also predict the structure in a sequential fashion as we do.\looseness=-1
Other work such as energy-based models \cite[\textit{inter alia}]{belanger2016structured,tu2018learning} and graphical models \cite{durrett-klein-2014-joint, ganea-hofmann-2017-deep} predict structures more holistically.\looseness=-1

%% file: tabs/result_test_conll.tex
\begin{table*}[!ht]
\centering \small
\resizebox{0.9\textwidth}{!}{
\begin{tabular}{lcccccccccc}
\toprule
                     & \multicolumn{3}{c}{MUC}     & \multicolumn{3}{c}{B$^3$}       & \multicolumn{3}{c}{CEAF$_{\phi_4}$}    &        \\ \cline{2-10} 
                   & P    & R    & F             & P    & R    & F             & P    & R    & F             &  Avg. F1        \\ \midrule
\citet{lee-etal-2017-end} & 78.4 & 73.4 & 75.8          & 68.6 & 61.8 & 65.0          & 62.7 & 59.0 & 60.8          & 67.2      \\ 
\citet{lee-etal-2018-higher} & 81.4 & 79.5 & 80.4          & 72.2 & 69.5 & 70.8          & 68.2 & 67.1 & 67.6          & 73.0      \\
\citet{joshi-etal-2019-bert} & 84.7 & 82.4 & 83.5          & 76.5 & 74.0 & 75.3          & 74.1 & 69.8 & 71.9          & 76.9      \\
 \citet{joshi-etal-2020-spanbert} & 85.8 & 84.8 & 85.3          & 78.3 & 77.9 & 78.1          & 76.4 & 74.2 & 75.3          & 79.6    \\
 \citeauthor{joshi-etal-2020-spanbert}$+\tfive_\textsc{b}$ $^\dagger$ \hspace{-0mm} & 82.4 & 77.4 & 79.8         & 72.3 & 68.2 & 70.2          & 70.5 & 63.5 & 66.8          & 72.3       \\ 
 \citeauthor{joshi-etal-2020-spanbert}$+\tfive_\textsc{l}$ $^\dagger$ & 85.5 & 77.7 & 81.4          & 78.3 & 68.5 & 73.1          & 75.0 & 65.9 & 73.1          & 74.9     \\ 
 \citeauthor{dobrovolskii-2021-word} & 84.9 & 87.9 & 86.3  & 77.4 & 82.6 &   79.9  & 76.1 & 77.1     & 76.6          & 81.0     \\ 
\midrule
\citeauthor{urbizu-etal-2020-sequence} & 
- & - & 64.9 &  - & - & 66.5 & - & - & 65.3 & 65.6   \\ 
 \citeauthor{tanl}$+\tfive_\textsc{b}$ & - & - & 81.0 &  - & - & 69.0 & - & - & 68.4 & 72.8   \\ 
\texttt{ASP}$+\tfive_\textsc{b}$ & 81.7  & 82.8 & 82.3 & 74.2 & 76.1 & 75.1 & 74.5 & 70.6 & 72.5 &  76.6  \\ 
\texttt{ASP}$+\tfive_\textsc{l}$ &  83.3 &  86.2 & 84.7 & 75.9 & 79.5 & 77.7 & 75.8 & 74.5 & 75.2 & 79.3  \\ 
\texttt{ASP}$+\flant_\textsc{l}$ & 83.5 & 87.6  & 85.5 & 76.3 & 81.8 & 79.0 & 76.0 & 76.2 & 76.1 & 80.2  \\ 
\texttt{ASP}$+\tzero_\textsc{3b}$ & 85.8  &  88.3 & 86.9 & 79.6 & 83.3 & 81.5 & 78.3 & 78.5 & 78.4 & 82.3   \\ 
\texttt{ASP}$+\flant_\textsc{xl}$ & 84.9  & 88.7 & 86.7 & 78.5 & 83.8 & 81.1 & 78.4 & 78.5 & 78.4 & 82.2 \\
\texttt{ASP}$+\flant_\textsc{xxl}$ & 86.1  & 88.4 & 87.2 & 80.2 & 83.2 & 81.7 & 78.9 & 78.3 & 78.6 & \textbf{82.5} \\ \bottomrule
\end{tabular}} 
\caption{Full results on the CoNLL-12 English test set. Avg. F1 denotes the average F1 of MUC, B{$^3$}, and CEAF{$_{\phi_4}$}. 
Models marked with ${}^\dagger$ are our re-implementation. Other results are taken from their original papers.}
\label{tab:result_test_conll}
\end{table*}